\documentclass[lettersize,journal]{IEEEtran}
\usepackage{amsmath,amsfonts}
\usepackage{algorithmic}
\usepackage{algorithm}
\usepackage{array}
\usepackage[caption=false,font=normalsize,labelfont=sf,textfont=sf]{subfig}
\usepackage{textcomp}
\usepackage{stfloats}
\usepackage{url}
\usepackage{verbatim}
\usepackage{graphicx}
\usepackage{cite}
\usepackage{amsmath}
\usepackage{amsthm}
\usepackage{booktabs}
\usepackage[switch]{lineno}
\usepackage{colortbl}
\usepackage{colortbl,xcolor}
\definecolor{lightgray}{gray}{0.9}        

\usepackage{amssymb}
\usepackage{multirow}
\usepackage{multicol}
\usepackage{arydshln}
\usepackage[driverfallback=dvipdfm,
CJKbookmarks=true, bookmarksnumbered=true, bookmarksopen=true,
colorlinks=true,
pdfborder=001,
citecolor=blue, linkcolor=blue, anchorcolor=red, urlcolor=black
]{hyperref}
\usepackage{cleveref}
\usepackage{mathtools}

\begin{document}

\title{One-Step Diffusion-based Real-World Image Super-Resolution with Visual Perception Distillation}

\author{Xue~Wu,
        Jingwei~Xin,
        Zhijun~Tu,
        Jie~Hu,
        Jie~Li,
        Nannan~Wang,~\IEEEmembership{Senior Member,~IEEE,}
        and~Xinbo~Gao,~\IEEEmembership{Fellow,~IEEE}
\thanks{Xue Wu, Jingwei Xin and Nannan Wang are with the State Key Laboratory of Integrated Services Networks, School of Telecommunications Engineering, Xidian University, Xi'an 710071, Shaanxi, China.
(e-mail: wuxue9898@stu.xidian.edu.cn; jwxin@xidian.edu.cn; nnwang@xidian.edu.cn.)}
\thanks{Zhijun Tu and Jie Hu are with Huawei Noah’s Ark Lab, Beijing 100084, China. (e-mail:zhijun.tu@huawei.com; hujie23@huawei.com)}
\thanks{Jie Li is with the State Key Laboratory of Integrated Services Networks, School of Electronic Engineering, Xidian University, Xi'an 710071, Shaanxi, China. (e-mail: leejie@mail.xidian.edu.cn)}
\thanks{Xinbo Gao is with the Chongqing Key Laboratory of Image Cognition, Chongqing University of Posts and Telecommunications, Chongqing 400065, China. (e-mail: gaoxb@cqupt.edu)}
}


\maketitle

\begin{abstract}
Diffusion-based models have been widely used in various visual generation tasks, showing promising results in image super-resolution (SR), while typically being limited by dozens or even hundreds of sampling steps. 
Although existing methods aim to accelerate the inference speed of multi-step diffusion-based SR methods through knowledge distillation, 
their generated images exhibit insufficient semantic alignment with real images, resulting in suboptimal perceptual quality reconstruction, specifically reflected in the CLIPIQA score. These methods still have many challenges in perceptual quality and semantic fidelity. Based on the challenges, we propose VPD-SR, a novel visual perception diffusion distillation framework specifically designed for SR, aiming to construct an effective and efficient one-step SR model. 
Specifically, VPD-SR consists of two components: Explicit Semantic-aware Supervision (ESS) and High-Frequency Perception (HFP) loss. Firstly, the ESS leverages the powerful visual perceptual understanding capabilities of the CLIP model to extract explicit semantic supervision, thereby enhancing semantic consistency. Then, Considering that high-frequency information contributes to the visual perception quality of images, in addition to the vanilla distillation loss, the HFP loss guides the student model to restore the missing high-frequency details in degraded images that are critical for enhancing perceptual quality.    
Lastly, we expand VPD-SR in adversarial training manner to further enhance the authenticity of the generated content.
Extensive experiments conducted on synthetic and real-world datasets demonstrate that the proposed VPD-SR achieves superior performance compared to both previous state-of-the-art methods and the teacher model with just one-step sampling.
\end{abstract}
\begin{IEEEkeywords}
Diffusion model, image super-resolution, one-step diffusion distillation, knowledge distillation.
\end{IEEEkeywords}

\section{Introduction}
\IEEEPARstart{S}{ingle} image super-resolution (SISR), a fundamental computer vision task, aims to reconstruct high-resolution (HR) images from their low-resolution (LR) counterparts affected by complex degradations~\cite{1survey}. SISR has found widespread applications across various domains, including medical imaging~\cite{56medical_TCSVT1,56medical_TCSVT2}, remote sensing~\cite{57remote_TMM,57remotesensing_TCSVT}, and so on~\cite{58video_tcsvt,Lu,xia2022cbash,Ye1}.
In past years, various generative adversarial network (GAN)-based~\cite{2gan,3ldl,4esrgan,5realesrgan} and Transformer-based SR approaches~\cite{7esrt,8IPT,rsa,freqformer} have been developed and shown remarkable power in the SISR task. Most recently, diffusion-based SR methods~\cite{10sr3,11srdiff,12resdiff} have surpassed previous techniques in terms of visual perceptual quality due to their superior ability to model complex data distributions, consequently attracting significant research attention.\\ 
\indent In SR tasks, current diffusion-based SR methods can be broadly classified into three categories. One modifies the inverse process of pre-trained diffusion models with gradient descent~\cite{18come,19sdedit}, another involves concatenating the LR image with the noise and training the diffusion models from scratch~\cite{11srdiff,13resshift}, and the third utilizes the robust generative prior of large-scale pre-trained Text-to-Image (T2I) diffusion models and adapts them to SR tasks by fine-tuning~\cite{20stablesr,21diffbir,22seesr, 23pasd}. Although achieving promising results, these methods typically require tens to hundreds of sampling steps to generate high-quality images, which significantly limits their practical application.\\
\begin{figure}[t]
  \centering
  \includegraphics[width=1\linewidth]{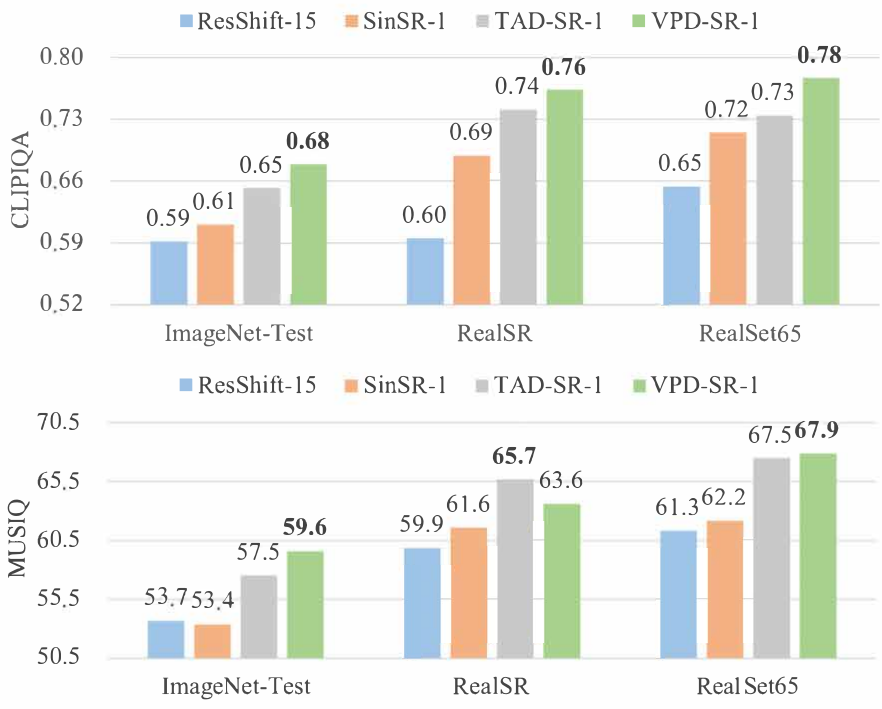}
  \centering
  \caption{Performance comparison on CLIPIQA (top) and MUSIQ (bottom) metrics of our proposed VPD-SR with other state-of-the-art methods across three datasets. Our method surpasses competitors on almost all datasets.}
  \label{fig1}
  \vspace{-0.5cm} 
\end{figure}
\indent To accelerate the generation process of diffusion models, various acceleration techniques have been proposed, such as the applications of numerical samplers~\cite{26ddim,29dpm++} and knowledge distillation~\cite{30progressive,32add}. However, directly applying these techniques to SR tasks poses significant challenges, as they often achieve acceleration at the cost of performance. 
More recently, 
ResShift~\cite{13resshift} established a shorter Markov chain and outperformed earlier diffusion-based SR approaches with only 15 sampling steps.
Furthermore, SinSR~\cite{14sinsr} and TAD-SR~\cite{15tadsr} condensed the inference steps of ResShift into a single step by distillation.
In addition, recent methods~\cite{33addsr,16osed,34s3r} adapted pre-trained T2I models into a one-step SR model by integrating adversarial distillation and distribution matching to accelerate inference. 
Despite these advancements, their performance in terms of the visual perceptual quality of recovered results remains insufficient (as indicated in the comparison of the CLIPIQA and MUSIQ scores in \Cref{fig1}. CLIPIQA leverages the pretrained CLIP model\cite{35clip} to evaluate semantic coherence and perceptual attributes within images by utilizing its cross-modal alignment capabilities. MUSIQ employs the hierarchical transformer to analyze perceptual quality across multiple granularities, enabling comprehensive assessment of both global composition and local details.), failing to enhance the semantic consistency between the generated images and ground truth (GT) images. 
Although some methods~\cite{59SSPIR,22seesr,23pasd} incorporated architectural and semantic information extracted from input LR images to produce realistic HR images, the severe degradation in the quality of LR images causes damage to local structures, resulting in semantic ambiguity in the images. This fundamental limitation prevents reliable extraction of semantic cues directly from real-world LR inputs. These methods lack explicit semantic supervision to guide the model to generate faithful semantic information, which is essential for diffusion models to effectively enhance semantic consistency.\\
\indent Furthermore, to explore more suitable distillation methods for accelerated diffusion-based SR models, we conducted an experiment to visualize step by step the the predictive outputs of the teacher model at each time step during inference, along with the corresponding low-frequency components, and polt their CLIPIQA scores trajectories, as shown in \Cref{fig:5}. From the \Cref{fig:5} (a) and (b), it can be seen that from large time steps to small time steps, the single-step output of the teacher model and the corresponding low-frequency components gradually become clearer and more realistic. In order to design an appropriate distillation strategy, we visualized the Fourier spectra of the prediction results at each time step in \Cref{fig:5} (a) and (b), respectively (as shown in \Cref{fig:5} (c) and (d), and further analyze the difference maps between the Fourier spectra at each time step (as shown in \Cref{fig:5} (e)). According to \Cref{fig:5} (c), (d), and (e), it can be seen that from the early to the late stages of the inference phase, the high-frequency components of the teacher model's predicted outputs continuously strengthen, while the low-frequency components remain stable. Furthermore, in the later stages, there is a significant increase in the high-frequency components. Additionally, as shown in \Cref{fig:51}, we plotted the CLIPIQA metric trajectories of both the teacher model's prediction output at each step and the corresponding low-frequency components. We found that the CLIPIQA scores of the prediction outputs at small time steps continuously increased while the CLIPIQA scores of the low-frequency components remained steady. Combined with the above findings, this indicates that the high-frequency information in the outputs of diffusion-based SR models contributes to improving the perceptual quality of the generated results.\\ 
\indent Motivated by the above two analyses, we propose a visual perception diffusion distillation method, termed VPD-SR, which exploits the explicit semantic-aware supervision from pre-trained CLIP model and high-frequency feature constraints to guide student moldel in generating semantically faithful and high-quality images, with the aim of developing an effective and efficient one-step SR model. Specifically, the explicit semantic-aware supervision improves semantic consistency and perceptual quality by integrating CLIP guidance and aligning generated images with GT images in the semantic space. To acquire semantic information that is more faithful to the original image, we extract the image embedding of the input image with CLIP image encoder.
Additionally, to ensure that the model produces images with fine detail while maintaining perceptual quality, in addition to the vanilla distillation loss, the high-frequency perception loss enforces  the generation of quality images by explicitly preserving high-frequency components, which demonstrate significant positive correlation with image visual perception metrics.
Finally, to further enhance the authenticity of the generated results, we incorporate a patch-based discriminator in adversarial training, guiding the student model to directly generate samples that lie on the manifold of real images in a single inference step. As shown in \Cref{fig1}, our method achieves superior visual quality on both synthetic and real-world datasets while requiring only one-step sampling, significantly outperforming state-of-the-art (SOTA) approaches.\\
\indent Overall, our main contributions are summarized as follows:
\begin{itemize}
    \item We propose an effective and efficient one-step diffusion distillation framework with explicit semantic-aware supervision and high-frequency perception constraint for the SR task, VPD-SR, to achieve high-quality restoration results.
    \item We introduce explicit semantic-aware supervision to provide faithful semantic guidance for the student model by exploiting the visual perceptual understanding capabilities of the CLIP. 
    \item To maintain high-frequency information that contributes to the visual perception quality of images, we employ a high-frequency perception loss to supervise high-frequency information in SR images.
    \item Although the distillation method involved in VPD-SR is straightforward, it achieves exceptional performance with just one sampling step, significantly outperforming previous SOTA methods on both synthetic and real-world datasets.
\end{itemize}
\begin{figure*}[!t]
  \centering
  \includegraphics[width=1\linewidth]{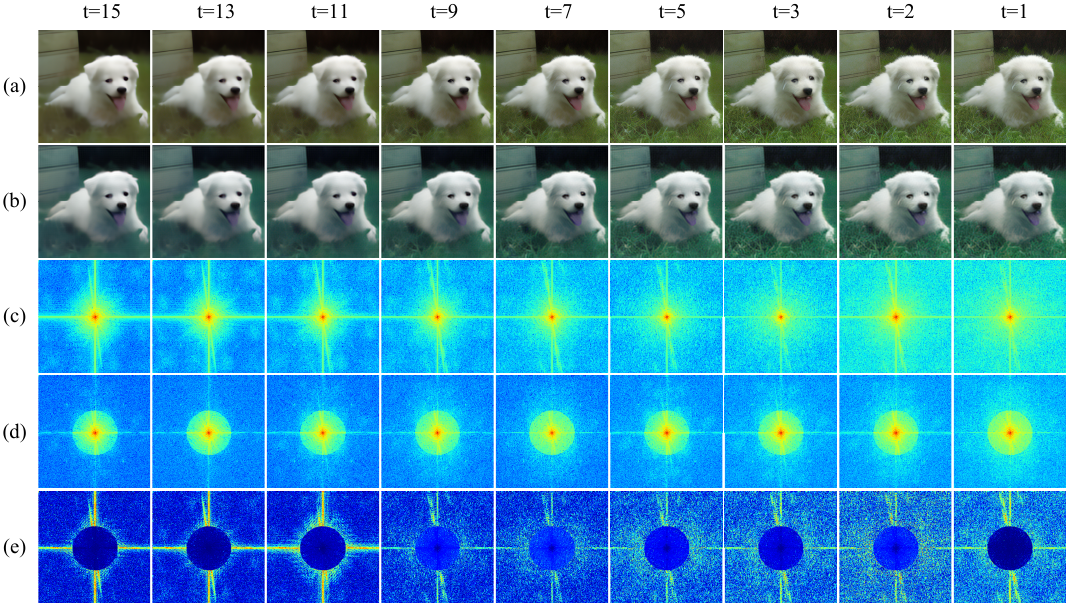}
  \centering
  \caption{Step-wise visualization of the teacher model (ResShift)'s one-step prediction during inference. (a) We visualize the predictive output of the teacher model at each sampling step. (b) Visualization of low-frequency components in the teacher model's predictions at each sampling step. (c) and (d) display the Fourier spectra of the predictions at each time step in (a) and (b), respectively. (e) represents the difference map between Fourier spectra at corresponding time points in (c) and (d). }
  \label{fig:5}
  \vspace{-0.5cm} 
\end{figure*}
\begin{figure}[!h]
  \centering
  \includegraphics[width=1\linewidth]{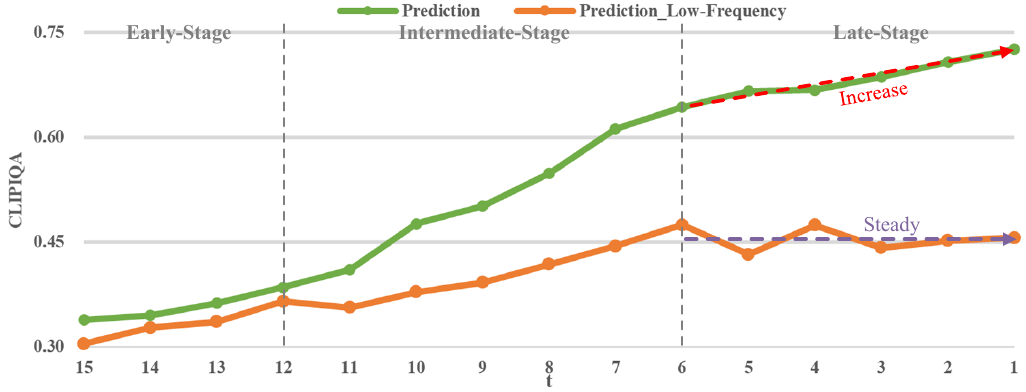}
  \centering
  \caption{The CLIPIQA score trajectories for both the teacher model's predictions and their low-frequency components at each time step.}
  \label{fig:51}
  \vspace{-0.5cm} 
\end{figure}
\vspace{-0.3cm} 
\section{Related Work}
\subsection{Image Super-Resolution}
Beginning with the introduction of convolutional neural network (CNNs) by SRCNN~\cite{41srcnn}, deep learning-based methods have gradually become the mainstream in the SR task. CNN-based SR methods~\cite{42multi} focus on improving fidelity. 
Although Transformer-based models~\cite{9swinir} have shown performance improvements, challenges such as unnatural artifacts and poor visual quality persist. GAN-based approaches~\cite{43realsrjpeg,44bsrgan}, while enhancing realism, often encounter issues like mode collapse and training instability. Recently, diffusion-based methods~\cite{45ldm,13resshift} for image SR tasks have garnered widespread attention due to their ability to model complex data distributions. For instance, the pioneering work SR3~\cite{10sr3}, which introduced diffusion models into the SR task, incorporated LR images as conditions to guide the generation of HR outputs.  
However, generating high-quality images directly from noise in pixel space may unnecessarily increase the burden on the diffusion model. To improve efficiency, some methods, such as StableSR~\cite{20stablesr} and DiffBIR~\cite{21diffbir}, leveraged the image priors of pre-trained stable diffusion (SD) model~\cite{45ldm} to generate high-quality SR outputs in latent space. Nonetheless, these methods typically require dozens of sampling steps, which limits their inference efficiency. 

\subsection{Acceleration of Diffusion Models}
Despite the remarkable advancements of diffusion models in image generation tasks in recent years, they typically require substantial sampling steps to produce high-quality results, which limits their feasibility for real-time generation and large-scale applications. 
To improve efficiency, researchers have proposed various acceleration strategies, such as numerical samplers and knowledge distillation. 
IDDPM~\cite{25iddpm} implemented a cosine noise schedule, which resulted in improved log-likelihood and facilitated faster sampling. 
DDIM~\cite{26ddim} generalized the Markovian forward diffusion of DDPMs~\cite{46ddpm} into non-Markovian processes and introduced a deterministic sampling method to accelerate the sampling process. Building upon DDIM, the DPM-solver~\cite{27dpmsolver} approximated the error prediction via a Taylor expansion, achieving efficient sampling by analytically resolving the linear component of the ordinary differential equation (ODE) solution. 
However, these methods condense the sampling steps at the cost of performance degradation and still require more than ten sampling steps to generate samples. In contrast, progressive distillation strategies~\cite{47distillation,30progressive} gradually reduce the inference steps of student models through multistage distillation, but they face the challenge of error accumulation. 
YONOS-SR~\cite{48YONOS-SR} uses knowledge distillation, instead of training faster samplers, it transfers knowledge from different scaling tasks and utilizes the DDIMs for efficient sampling. Additionally, adversarial diffusion distillation (ADD)~\cite{32add} combines an adversarial objective with a score distillation objective to distill the pre-trained SD models. AddSR~\cite{33addsr} employs ADD for SR task, resulting in a comparatively effective four-step model. OSEDif~\cite{16osed} directly takes the given LQ image as the starting point for diffusion and performs the variational score distillation (VSD)~\cite{31prolificdreamer} in the latent space. ResShift~\cite{13resshift} was proposed to enhance the inference efficiency, which builds up a shorter Markov chain between the LR and HR images and performs better than the earlier diffusion-based SR approaches with only 15 sampling steps during inference. SinSR~\cite{14sinsr} proposes a one-step diffusion model by distilling the ResShift method with consistency preserving distillation. TAD-SR~\cite{15tadsr} introduces a high-frequency enhanced score distillation and designs a time-aware discriminator to further enhance the performance of the student model. 
\begin{figure*}[!t]
  \centering
  \includegraphics[width=1\linewidth]{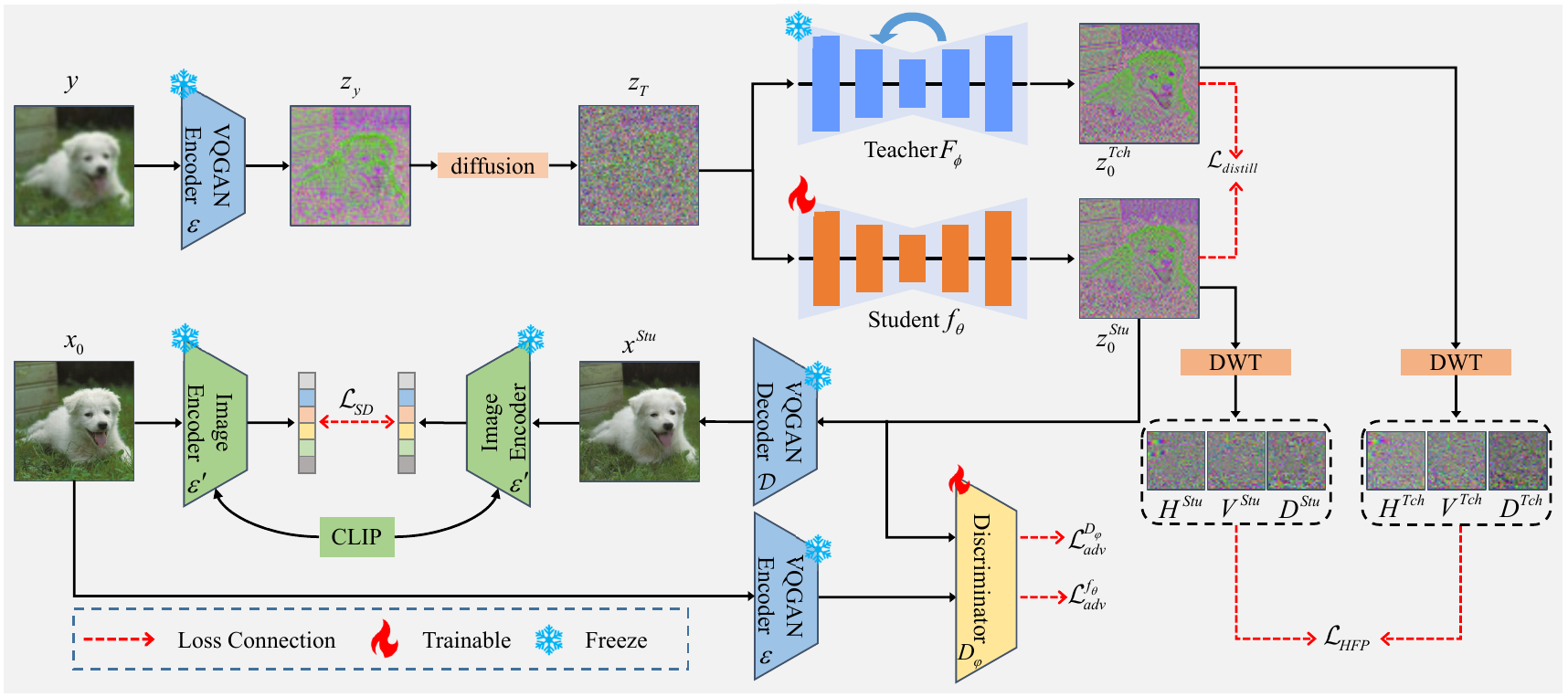}
  \centering
  \caption{Overall framework of our proposed VPD-SR. We distill the given pre-trained teacher model into a fast one-step student model. To match the single-step outputs of the student model with the multi-step sampling results of the teacher model, we optimize the student model using the proposed high-frequency perception loss, semantic distillation loss, and distillation loss. Additionally, we incorporate generative adversarial learning into the diffusion distillation framework to enhance the authenticity of the generated results.}
  \label{fig:2}
  \vspace{-0.5cm} 
\end{figure*}
\section{Methodology}
\subsection{Preliminary}
The diffusion model is a generative model that leverages the diffusion process to model the data distribution. In SR tasks, given a LR image $y$ and its corresponding HR image $x_0$, general diffusion-based SR methods gradually add Gaussian noise to the $x_0$ through a Markov chain where the forward process is typically  defined as $q\left( {{x_t}|{x_{t - 1}}} \right) = {\cal N}\left( {{x_t};\sqrt {1 - {\beta _t}} {x_{t - 1}},{\beta _t}I} \right)$ with an initial state ${x_T}\sim{\cal N}\left( {0,I} \right)$, and iteratively recover $x_0$ from pure noise through a reverse process conditioned on $y$, denoted ${p_\theta }\left( {{x_{t - 1}}|{x_t},y} \right)$, $\theta$ represents the parameters of the deep network used for learning the reverse process. ResShift shortens the Markov chain by incorporating the information of the LR image $y$ into the initial state $x_T$, where the forward process of the Markov chain can be formulated as follows:
\begin{equation}
q\left( {{x_t}|{x_0},y} \right) = {\cal N}\left( {{x_t};{x_0} + {\eta _t}\left( {y - {x_0}} \right),{k^2}{\eta _t}I} \right),
\label{eq:1}
\end{equation}
where $\eta _t$ is a serial of hyper-parameters that monotonically increases with timestep $t$ and satisfies ${\eta _0} \to 0$ and ${\eta _T} \to 1$, $k$ is a hyper-parameter controlling the noise variance. The corresponding inverse diffusion process starts from the initial state ${x_T} = y + k\sqrt {{\eta _T}}{\epsilon}$ where ${\epsilon}\sim{\cal N}\left( {0,I} \right)$, which can be expressed as follows:
\begin{equation}
q\left( {{x_{t - 1}}|{x_t},{x_0},y} \right) = {\cal N}( {{x_{t - 1}};\frac{{{\eta _{t - 1}}}}{{{\eta _t}}}{x_t} + \frac{{{\alpha _t}}}{{{\eta _t}}}{x_0},{k^2}\frac{{{\eta _{t - 1}}}}{{{\eta _t}}}{\alpha _t}I}),
\label{eq:2}
\end{equation}
where ${\alpha _t} = {\eta _t} - {\eta _{t - 1}}$. To mitigate the randomness of the generated image due to the existence of the random noise in initial state $x_T$, SinSR~\cite{14sinsr} reformulates \cref{eq:2} by employing deterministic sampling as follows:
\begin{equation}
q\left( {{x_{t - 1}}|{x_t},{x_0},y} \right) = \delta \left( {{k_t}{x_0} + {m_t}{x_t} + {j_t}y} \right),
\end{equation}
where $\delta$ is the unit impulse, ${m_t} = \sqrt {{\textstyle{{{\eta _{t - 1}}} \over {{\eta _t}}}}}$, ${j_t} = {\eta _{t - 1}} - \sqrt {{\eta _{t - 1}}{\eta _t}}$, and ${k_t} = 1 - {m_t} - {j_t}$. The details of the derivation is described in the SinSR. During inference, $x_{t-1}$ can be calculated from $x_t$ using following equation:
\begin{equation}
{x_{t - 1}} = {k_t}{{\hat x}_0} + {m_t}{x_t} + {j_t}y
\end{equation}
where ${\hat x}_0$ is usually predicted by a trainable deep network ${f_\theta }\left({{x_t},y,t} \right)$ with parameter $\theta$. The training objective function of $f_\theta$ as follows:
\begin{equation}
\mathop {\min }\limits_\theta  \sum\nolimits_t {{w_t}} \left\| {{f_\theta }\left( {{x_t},y,t} \right) - {x_0}} \right\|_2^2
\end{equation}
where ${w_t} = {\textstyle{{{\alpha _t}} \over {2{k^2}{\eta _{t - 1}}{\eta _t}}}}$. In practice, the omission of weight $w_t$ results in performance improvement~\cite{46ddpm}. Following SinSR, we utilized the pre-trained ResShift model using deterministic sampling as our teacher model.
\subsection{Overview of VPD-SR}
As illustrated in \Cref{fig:2}, our goal is to distill a given pre-trained teacher model into a fast one-step student model. The overall framework of our proposed VPD-SR consists a teacher model $F_\phi$, a student model $f_\theta$ initialized from the teacher model, a trainable discriminator $D_\varphi$, a encoder $\varepsilon$ and a decoder $\cal D$ of VQGAN~\cite{49vqgan}, and a image encoder $\varepsilon '$ of CLIP model. Given a low-resolution input $y$, the VQGAN encoder encodes $y$ into a latent code ${z_y} = \varepsilon \left( y \right)$ and the initial state ${z_T} = {z_y} + k\sqrt {{\eta _T}} \epsilon$ is obtained through the forward process \cref{eq:1}. The $z_T$ is then processed by the teacher model and the student model to generate the enhanced latent codes $z_0^{Tch}$ and $z_0^{Stu}$, respectively. Finally, the VQGAN decoder reconstructs the HR output $x^{Stu}$. During training, we introduce explicit semantic-aware supervision to improve the semantic consistency and perceptual quality of the generated images. In addition, we compute the vanilla distillation loss and the high-frequency perception loss between the outputs of the teacher and student models in the latent space. Furthermore, we utilize a patch-based discriminator~\cite{50patch} to enhance the authenticity of the generated content through adversarial training.
\subsection{Explicit Semantic-aware Supervision}
To address the limitations of traditional loss functions and better align the generated images with visual perception, we propose explicit semantic-aware supervision, which extracts semantic information from the GT image through the image encoder of a pretrained CLIP model as semantic supervision, and introduce aligning the embeddings of SR and GT images in semantic space to enhance visual perception quality, rather than relying on semantic guidance like other models~\cite{59SSPIR,22seesr,23pasd}. Note that the pretrained CLIP model, which is trained on a dataset of 4 billion image-text pairs, possesses powerful cross-modal alignment capabilities, enabling robust visual-semantic 
understanding.\\
\indent Explicit semantic-aware supervision enforces semantic consistency by aligning the generated images with GT images in semantic-level. For the output of student model $x^{Stu}$ and its GT image $x_0$, we compute their embedding vectors through CLIP image encoder $\varepsilon '$ as follows:
\begin{equation}
{v_{GT}} = \varepsilon '\left( {{x_0}} \right),{v_{SR}} = \varepsilon '\left( {{x^{Stu}}} \right),
\end{equation}
where $v_{GT}$ and $v_{SR}$ represent the embedding vectors of the GT image $x_0$ and the output of student model $x^{Stu}$ in the semantic space, respectively. To assess semantic alignment, we calculate the cosine similarity between $v_{GT}$ and $v_{SR}$ as follows:
\begin{equation}
cos = \frac{{\left\langle {{v_{SR}},{v_{GT}}} \right\rangle }}{{\left\| {{v_{SR}}} \right\| \cdot \left\| {{v_{GT}}} \right\|}},
\end{equation}
where $\cos  \in \left[ { - 1,1} \right]$ denotes the degree of semantic alignment with the value closer to 1 indicating that the two vectors are more similar, $\left\langle { \cdot , \cdot } \right\rangle$ represents the inner product of vectors. The degree of semantic alignment quantifies the semantic consistency between the restored image and its GT image. The semantic distillation loss is then defined as:
\begin{equation}
{{\cal L}_{SD}} = 1 - cos ,
\end{equation}
which encourages the generated image to preserve more of the semantic content of the GT image.

\subsection{High-Frequency Perception Loss}
In image SR task, high-frequency (HF) detail features are crucial because downgraded images lack detailed information. In addition, high-frequency information contributes to the visual perception quality of images. The main goal of SR is to recover the missing information as much as possible from the degraded image while ensuring that the recovered images are of high perceptual quality. In addition to the vanilla distillation loss, We introduce a high-frequency perception loss ${{\cal L}_{HFP}}$ based on the Discrete Wavelet Transform (DWT) to preserve the HF feature. \\
\indent Performing DWT on an image decomposes it into four sub-bands: low-low (LL), low-high (LH), high-low (HL), and high-high (HH). The LL sub-band captures the low-frequency content of the image, while the remaining three sub-bands represent the high-frequency components of the image in the horizontal, vertical, and diagonal directions, respectively. Given the output $z_0^{Tch}$ of teacher model and the output $z_0^{Stu}$ of student model, we extract the wavelet coefficients of the high-frequency bands $H$, $V$, and $D$, which refer to the high-frequency components in the horizontal, vertical, and diagonal directions, respectively, through DWT operation. We improve detailed information and perceptual quality by utilizing high-frequency perception loss, which can be expressed as follows:
\begin{align}
    {{\cal L}_{HFP}} = & \mathbb{E}[{{\left\| {{H^{Tch}} - {H^{Stu}}} \right\|}^2} +
                        {{\left\| {{V^{Tch}} - {V^{Stu}}} \right\|}^2}  \nonumber \\
                    + & {{\left\| {{D^{Tch}} - {D^{Stu}}} \right\|}^2}] 
\end{align}
where $H^{Tch}$, $V^{Tch}$, and $D^{Tch}$ are the sub-bands of the teacher model output. $z_0^{Tch}$, $H^{Stu}$, $V^{Stu}$, and $D^{Stu}$ are the sub-bands of the student model output $z_0^{Stu}$.\\
\indent The vanilla distillation loss is formulated as follows:
\begin{equation}
{{\cal L}_{distill}} = {{\cal L}_{MSE}}\left( {{F_\phi }\left( {{z_T},T,y} \right),{f_\theta }\left( {{z_T},T,y} \right)} \right),
\end{equation}
which guides the student model to establish the mapping between the LR and HR images in just one step by utilizing the T-step outputs of the teacher model as the learning objective for the student model.
\vspace{-0.3cm} 
\subsection{Comprehensive Training Objective}
Recent study~\cite{32add} has shown that combining diffusion models with generative adversarial networks can significantly improve the perceptual quality of generated images. Therefore, in addition to the aforementioned loss functions, we also integrate the adversarial loss into the diffusion model to enhance quality of generated images. Specifically, we first utilize the VQGAN encoder to encode GT image $x_0$ into a latent code ${z_0} = \varepsilon \left( x_0 \right)$, then the real code $z_0$ and generated code $z_0^{Stu}$ are sent to a patch-based discriminator $D_\varphi$ that aims to differentiate between real and reconstructed contents. This discriminator evaluates each patch within an image to determine whether it is real or fake, then aggregates all individual responses to generate the final output. The detailed configuration of the discriminator can be found in ~\cite{50patch}. The corresponding adversarial loss can be formulated as follows:
\begin{equation}
{\cal L}_{adv}^{{f_\theta }} =  - {\mathbb{E}_{z_0^{Stu}}}[{D_\varphi }\left( {z_0^{Stu}} \right)]
\end{equation}
\begin{align}
{\cal L}_{adv}^{{D_\varphi }} = & {\mathbb{E}_{z_0}}[\max \left( {0,1 - {D_\varphi }\left( {{z_0}} \right)} \right)] \nonumber \\
+ & {\mathbb{E}_{z_0^{Stu}}}[\max \left( {0,1 + {D_\varphi }\left( {z_0^{Stu}} \right)} \right)]
\end{align}
where ${\cal L}_{adv}^{f_\theta }$ and ${\cal L}_{adv}^{{D_\varphi }}$ are applied to optimize the student model and the discriminator, respectively.\\
\indent Therefore, the overall training objective for the student model $f_\theta$ is:
\begin{equation}
{{\cal L}_{{f_\theta }}} = {{\cal L}_{distill}} + {\lambda _1}{{\cal L}_{HFP}} + {\lambda _2}{{\cal L}_{SD}} + {\lambda _3}{\cal L}_{adv}^{{f_\theta }}
\end{equation}
where $\lambda _1$, $\lambda _2$, and $\lambda _3$ are hyper-parameters to control the relative importance of these objectives. The overall of the
proposed method is summarized in \Cref{alg:algorithm}. 
\begin{algorithm}[h]
    \caption{Training VPD-SR}
    \label{alg:algorithm}
    \textbf{Require}: Pre-trained T-step teacher model $F_\phi$, student model $f_\theta$, trainable discriminator $D_\varphi$, frozen VQGAN encoder $\varepsilon$ and decoder $\cal D$, and frozen CLIP image encoder $\varepsilon '$\\
    \textbf{Require}: Paired training set $\left( {X,Y} \right)$, where $X$ and $Y$ represent the ground-truth and low-resolution image sets, respectively
    \begin{algorithmic}[1] 
        \STATE Initialize student model $f_\theta$ with teacher model $F_\phi$
        \WHILE{not converged}
        \STATE sample ${x_0,y}\sim\left( {X,Y} \right)$, sample ${\epsilon}\sim{\cal N}\left( {0,I} \right)$
        \STATE ${z_y} = \varepsilon \left( y \right)$, ${z_0} = \varepsilon \left( x_0 \right)$, ${z_T} = {z_y} + k\sqrt {{\eta _T}} \epsilon$
        \STATE $z_{0}^{Tch}={F_\phi }\left( {{z_T},T,y} \right)$, $z_{0}^{Stu}={f_\theta}\left( {{z_T},T,y} \right)$
        \STATE ${{\cal L}_{distill}} = {{\cal L}_{MSE}}\left( {z_{0}^{Tch},z_{0}^{Stu}} \right)$
        \STATE ${{\cal L}_{HFP}}=\mathbb{E}[{{\cal L}_{MSE}}\left(DWT(z_{0}^{Tch}),DWT(z_{0}^{Stu})\right)]$
        \STATE ${v_{GT}} = \varepsilon '\left( {{x_0}} \right), {x^{Stu}}={\cal D}(z_{0}^{Stu}), {v_{SR}} = \varepsilon '\left( {{x^{Stu}}} \right)$
        \STATE ${{\cal L}_{SD}}=1-\frac{{\left\langle {{v_{SR}},{v_{GT}}} \right\rangle }}{{\left\| {{v_{SR}}} \right\| \cdot \left\| {{v_{GT}}} \right\|}}$
        \STATE ${\cal L}_{adv}^{{f_\theta }} =  - {\mathbb{E}_{z_0^{Stu}}}[{D_\varphi }\left( {z_0^{Stu}} \right)]$
        \STATE ${{\cal L}_{{f_\theta }}} = {{\cal L}_{distill}} + {\lambda _1}{{\cal L}_{HFP}} + {\lambda _2}{{\cal L}_{SD}} + {\lambda _3}{\cal L}_{adv}^{{f_\theta }}$
        \STATE $f_\theta \leftarrow update\left(f_\theta,{{\cal L}_{{f_\theta }}}\right)$
        \STATE $  \begin{multlined}[t]
        {\cal L}_{adv}^{{D_\varphi }} = {\mathbb{E}_{z_0}}[\max \left( {0,1 - {D_\varphi                                   }\left( {{z_0}} \right)} \right)] \\ 
                                        \quad \quad + {\mathbb{E}_{z_0^{Stu}}}[\max \left( {0,1 + {D_\varphi }\left( {z_0^{Stu}} \right)} \right)]
                \end{multlined} $
        \STATE ${D_\varphi} \leftarrow update\left({D_\varphi},{\cal L}_{adv}^{{D_\varphi }}\right)$
        \ENDWHILE
        \STATE \textbf{return} The student model $f_\theta$
    \end{algorithmic}
\end{algorithm}
\begin{table*}[!t]
    \caption{Quantitative comparison of different SR methods on the $ImageNet$-$Test$ dataset. The best and the second best results are highlighted in \textbf{bold} and \underline{underline}, respectively.Running time is tested on NVIDIA Telsa A100 GPU on the $\times 4$ ($64\rightarrow256$) SR tasks. \# indicates trainable parameters. \label{tab:1}}
    \centering
        \renewcommand{\arraystretch}{1.2} 
        \renewcommand\tabcolsep{18pt}
        \begin{tabular}{r|ccccc}
            \hline
            \multirow{2}{*}{Methods} & \multicolumn{5}{c}{Metrics}  \\
            \cline{2-6}
            & LPIPS$\downarrow$ & CLIPIQA$\uparrow$ & MUSIQ$\uparrow$  &Runtime (s) & \# Params (M)\\
            \hline
            ESRGAN~\cite{4esrgan}      
            & 0.485 & 0.451   & 43.615 & 0.038 &16.70 \\
            RealSR-JPEG~\cite{43realsrjpeg} 
            & 0.326 & 0.537   & 46.981 & 0.038 &16.70\\
            BSRGAN~\cite{44bsrgan}      
            & 0.259 & 0.581   & 54.697 & 0.038 &16.70\\
            SwinIR~\cite{9swinir}      
            & 0.238 & 0.564   & 53.790 & 0.107 &28.01\\
            RealESRGAN~\cite{5realesrgan}  
            & 0.254 & 0.523   & 52.538 & 0.038 &16.70\\
            DASR~\cite{54dasr}        
            & 0.250 & 0.536   & 48.337 & 0.022 &8.06\\
            \hline
            LDM-15~\cite{45ldm}      
            & 0.269 & 0.512   & 46.419 & 0.408 &113.60\\
            ResShift-15~\cite{13resshift}    
            & 0.231 & 0.592   & 53.660 & 0.682 &118.59\\
            \hline
            SinSR-1~\cite{14sinsr}       
            & \textbf{0.221} & 0.611   & 53.357 & 0.058 &118.59\\
            TAD-SR-1~\cite{15tadsr}      
            & 0.227 & \underline{0.652}   & \underline{57.533} & 0.058 &118.59\\
            \rowcolor{lightgray}  
            VPD-SR-1      
            & \underline{0.226} & \textbf{0.683}   & \textbf{59.570} & 0.058 &118.59\\
            \hline
        \end{tabular}
        \vspace{-0.3cm} 
\end{table*}
\section{Experiments}
\subsection{Experimental Setup}
\textbf{Training and Testing Datasets.} We train the models on the training set of ImageNet~\cite{51imagenet} by following the same procedure as ResShift~\cite{13resshift} and SinSR~\cite{14sinsr} where the degradation pipeline of Real-ESRGAN~\cite{5realesrgan} is used to synthesize HR-LR pairs. We evaluate our model on synthetic dataset ImageNet-Test~\cite{13resshift} and three real-word datasets, RealSR~\cite{52realsr}, RealSet65~\cite{13resshift}, and DRealSR~\cite{53drealsr}.\\
\indent \textbf{Evaluation Metrics.} For evaluating the performance of various methods, we employ LPIPS~\cite{40lpips} and no-reference metrics (include CLIPIQA~\cite{36clipiqa}, MUSIQ~\cite{37musiq}, MANIQA~\cite{39maniqa}, and NIQE~\cite{38niqe}). 
Note that we take non-reference metrics as the primary evaluation metrics on the real-world datasets as they are more closely aligned with human subjective evaluations and perception.\\
\indent \textbf{Implementation Details.} For a fair comparison, we follow the same backbone design and parameter setup as that in ResShift and SinSR. Specifically, we initialize the student model with the teacher model, $i.e.$, ResShift. The weights of the losses $\lambda _1$, $\lambda _2$, and $\lambda _3$ are set to 0.1, 1, and 0.1, respectively. We only train student model, frozen the encoder and decoder of VQGAN and the image encoder of CLIP model in order to preserve their priors. We train the proposed VPD-SR for 30$K$ iterations with a batch size of 16 based on our proposed loss functions with a NVIDIA Tesla V100 GPU.\\
\indent \textbf{Compared methods.} We compare the performance pf our proposed method with several representative SR models, including ESRGAN~\cite{4esrgan}, RealSR-JPEG~\cite{43realsrjpeg}, BSRGAN~\cite{44bsrgan}, SwinIR~\cite{9swinir}, RealESRGAN~\cite{5realesrgan}, DASR~\cite{54dasr}, LDM~\cite{45ldm}, ResShift~\cite{13resshift}, SinSR~\cite{14sinsr}, TAD-SR\cite{15tadsr}, OSEDiff~\cite{16osed}, SSP-IR~\cite{59SSPIR}, StableSR~\cite{20stablesr}, DiffBIR~\cite{21diffbir}, SeeSR~\cite{22seesr}, and PASD~\cite{23pasd}.
\\
\vspace{-0.5cm} 

\subsection{Comparison with State-of-the-Arts}
\textbf{Quantitative comparisons on synthetic dataset.}
We conduct the quantitative comparison between the proposed VPD-SR and other state-of-the-art SR methods on the ImageNet-Test dataset, and the comparison results are reported in \Cref{tab:1}. As shown in the table, the proposed VPD-SR achieves the best CLIPIQA and MUSIQ scores among widely used GAN, Transformer, and diffusion model-based SR algorithms. Higher CLIPIQA and MUSIQ scores indicate that the generated image has better visual quality, such as semantic consistency and visual coherence. The substantial improvement in non-reference metrics indicates that VPD-SR has the ability to generate images with high perceptual quality and realism. It demonstrates the effectiveness of the proposed method. Furthermore, our VPD-SR achieves the second-best LPIPS score among the ten other SR models, outperforming the teacher model ResShift. \\
\begin{table*}[!t]
    \caption{Quantitative comparison among different SR methods on two real-word datasets. * indicates that the RealSR dataset is same as that in ResShift and SinSR. The best and the second best results are highlighted in \textbf{bold} and \underline{underline}, respectively.}
    \label{tab:2}
    \centering
        \renewcommand{\arraystretch}{1.2} 
        \renewcommand\tabcolsep{26pt}
        \begin{tabular}{r|cc|cc}
            \hline
            \multirow{3}{*}{Methods} & \multicolumn{4}{c}{Datasets} \\ 
            \cline{2-5} 
            & \multicolumn{2}{c|}{RealSR*} & \multicolumn{2}{c}{RealSet65} \\ 
            \cline{2-5} 
            & CLIPIQA$\uparrow$ & MUSIQ$\uparrow$  & CLIPIQA$\uparrow$ & MUSIQ$\uparrow$     \\ 
            \hline
            ESRGAN~\cite{4esrgan}                  & 0.2362       & 29.048      & 0.3739         & 42.369       \\
            RealSR-JPEG~\cite{43realsrjpeg}             & 0.3615       & 36.076      & 0.5282         & 50.539       \\
            BSRGAN~\cite{44bsrgan}                  & 0.5439       & 63.586      & 0.6163         & 65.582       \\
            SwinIR~\cite{9swinir}                  & 0.4654       & 59.636      & 0.5782         & 63.822       \\
            RealESRGAN~\cite{5realesrgan}              & 0.4898       & 59.678      & 0.5995         & 63.220       \\
            DASR~\cite{54dasr}                    & 0.3629       & 45.825      & 0.4965         & 55.708       \\
            \hline
            LDM-15~\cite{45ldm}                  & 0.3836       & 49.317      & 0.4274         & 47.488       \\
            ResShift-15~\cite{13resshift}             & 0.5958       & 59.873      & 0.6537         & 61.330       \\
            \hline
            SinSR-1~\cite{14sinsr}                 & 0.6887       & 61.582      & 0.7150         & 62.169       \\
            TAD-SR-1~\cite{15tadsr}                & \underline{0.7410} & \textbf{65.701} & \underline{0.7340}                                 & \underline{67.500}       \\
            \rowcolor{lightgray}  
            VPD-SR-1                & \textbf{0.7635} & \underline{63.592} & \textbf{0.7817}         
                                    & \textbf{67.918}       \\ 
            \hline
        \end{tabular}
        \vspace{-0.3cm} 
\end{table*}
\begin{table*}[!ht]
    \caption{Quantitative comparison among different SR methods on two real-word datasets. $\dagger$ indicates that the RealSR dataset is same as that in OSEDiff. The best and the second best results are highlighted in \textbf{bold} and \underline{underline}, respectively.}
    \label{tab:3}
    \centering
    \renewcommand{\arraystretch}{1.2} 
    \renewcommand\tabcolsep{8pt}
    \begin{tabular}{r|cccc|cccc}
        \hline
        \multirow{3}{*}{Methods} & \multicolumn{8}{c}{Datasets} \\ 
        \cline{2-9} 
        & \multicolumn{4}{c|}{RealSR$\dagger$} & \multicolumn{4}{c}{DRealSR} \\ 
        \cline{2-9} 
        & CLIPIQA$\uparrow$ & MUSIQ$\uparrow$  & MANIQA$\uparrow$ & NIQE$\downarrow$ 
        & CLIPIQA$\uparrow$ & MUSIQ$\uparrow$  & MANIQA$\uparrow$ & NIQE$\downarrow$    \\ 
        \hline
        StableSR-200~\cite{20stablesr} & 0.6178  & 65.78 & 0.6221 & 5.9122 & 0.6356  & 58.51 & 0.5601 & 6.5239 \\
        DiffBIR-50~\cite{21diffbir}   & 0.6463  & 64.98 & 0.6246 & 5.5346 & 0.6395  & 61.07 & 0.5930 & 6.3124 \\
        SeeSR-50~\cite{22seesr}     & 0.6612  & \underline{69.77} & \underline{0.6442} & \textbf{5.4081} & 0.6804  & \textbf{64.93} & \underline{0.6042} & 6.3967 \\
        PASD-20~\cite{23pasd}      & 0.6620  & 68.75 & \textbf{0.6487} & \underline{5.4137} & 0.6808  & 64.87 & \textbf{0.6169} & \textbf{5.5474} \\
        SSP-IR-50~\cite{59SSPIR}      & \underline{0.6765}  & \textbf{69.81} & 0.5544 & 5.5711 & 0.6955  & \underline{66.27} & 0.5255 & 6.6904 \\
        ResShift-15~\cite{13resshift}  & 0.5444  & 58.43 & 0.5285 & 7.2635 & 0.5342  & 50.60 & 0.4586 & 8.1249 \\
        \hline
        SinSR-1~\cite{14sinsr}      & 0.6122  & 60.80 & 0.5385 & 6.2872 & 0.6383  & 55.33 & 0.4884 & 6.9907 \\
        OSEDiff-1~\cite{16osed}    & 0.6693  & 69.09 & 0.6326 & 5.6476 & \underline{0.6963}  & 64.65 & 0.5899 & 6.4902 \\
        \rowcolor{lightgray}  
        VPD-SR-1     & \textbf{0.7322}  & 66.38 & 0.5817 & 5.7508 & \textbf{0.7272}  & 62.35 & 0.5323 & \underline{5.9779} \\ 
        \hline
    \end{tabular}
    \vspace{-0.3cm} 
\end{table*}
\textbf{Quantitative comparisons on real-world datasets.} 
In addition to evaluating our method on synthetic dataset, we also quantitatively compare the VPD-SR with other SR methods on four real-world datasets. The quantitative comparisons among various SR methods are presented in \Cref{tab:2} and \Cref{tab:3}. \Cref{tab:2} lists the comparative evaluation using CLIPIQA and MUSIQ of various methods on the RealSR* and RealSet65 datasets. 
As shown in the \Cref{tab:2}, in terms of the CLIPIQA metric, our proposed method significantly outperforms the ten other SR methods by a substantial margin on both RealSR* and RealSet65 datasets. 
Specifically, VPD-SR exceeds the teacher model ResShift with just a single inference step by 28.1\% and 19.6\% and surpasses the second-best method with a notable advantage in CLIPIQA scores on the RealSR* and RealSet65 datasets, respectively. In terms of the MUSIQ metric, our proposed method achieves the best result on the RealSet65 dataset and the second-best score on the RealSR* dataset. These demonstrate the validity of our proposed method. In addition to comparing our method with the single-step diffusion models, we also compare it with the multi-step diffusion models by using CLIPIQA, MUSIQ, MANIQA, and NIQE metrics, on the RealSR$\dagger$ and DRealSR datasets. Note that the real-world data from both datasets include LQ images of size 128 $\times$ 128 pixels. From \Cref{tab:3}, VPD-SR demonstrates significant advantages over competing methods in terms of CLIPIQA metric and achieves competitive results in terms of NIQE metric on the both datasets. 
However, in MUSIQ and MANIQA metrics, the SeeSR and PASD exhibit better performance, which may be attributed to the fact that multi-step models have more denoising iterations to produce rich details than a single-step inference. The above results indicate the ability of our proposed method to generate results with high visual perceptual quality.\\ 
\begin{figure*}[!t]
  \centering
  \includegraphics[width=1\linewidth]{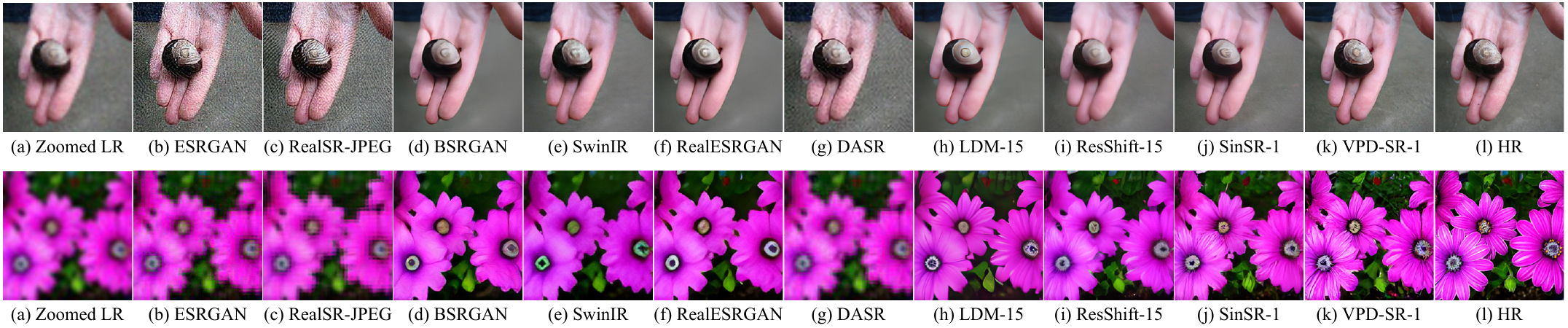}
  \centering
  \caption{Qualitative comparisons of different methods on two synthetic examples of the $ImageNet$-$Test$ dataset. Please zoom in for a better view.}
  \label{fig:3}
\end{figure*}
\begin{figure*}[!h]
  \centering
  \includegraphics[width=1\linewidth]{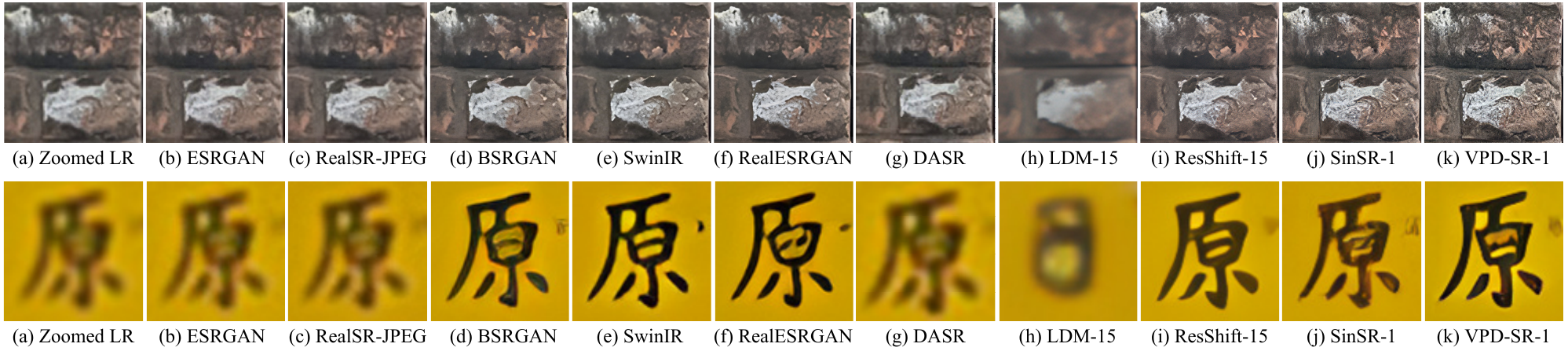}
  \centering
  \caption{Qualitative comparisons of different methods on two real examples of the $RealSR$* and $RealSet65$ datasets. Please zoom in for a better view.}
  \label{fig:4}
\end{figure*}
\begin{figure*}[!htb]
  \centering
  \includegraphics[width=1\linewidth]{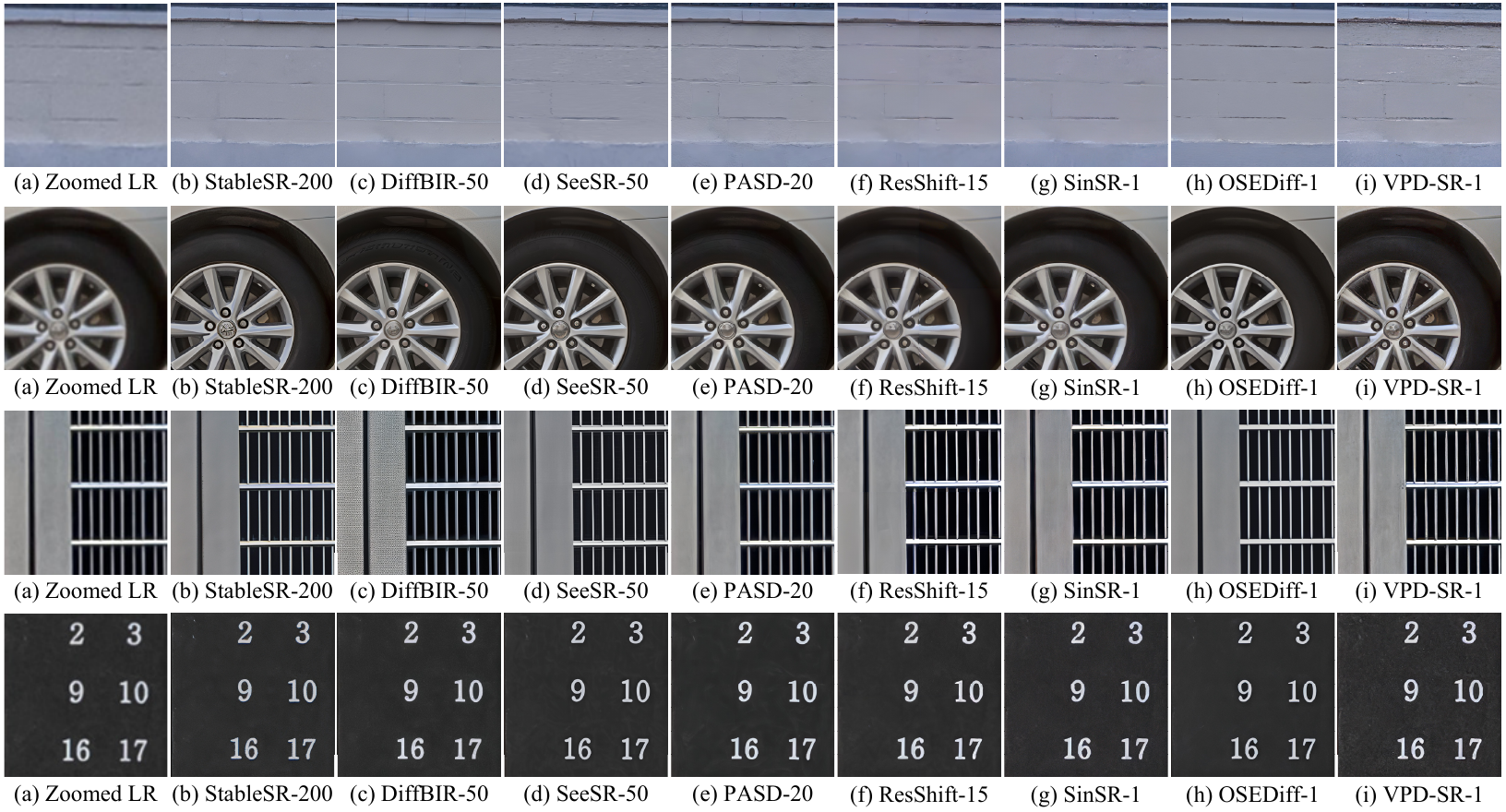}
  \centering
  \caption{Qualitative comparisons of different methods on four real examples of the $RealSR\dagger$ and $DRealSR$ datasets. Please zoom in for a better view.}
  \label{fig:6}
  \vspace{-0.5cm} 
\end{figure*}
\textbf{Qualitative Comparisons.} We present qualitative comparisons between our proposed method and other leading SR models in \Cref{fig:3,fig:4,fig:6}. TAD-SR~\cite{15tadsr} and SSP-IR~\cite{59SSPIR} are not included since their official code is unavailable. Visual results tested on both synthetic and real-world datasets show that VPD-SR produces images with higher clarity and better visual perception. To ensure a comprehensive evaluation, we compared various scenarios, such as buildings, text, and natural landscapes. It can be observed that the images generated by VPD-SR are more natural, which is evident from the distinct wall and window textures (as shown in the 1st sample of \Cref{fig:4} and the 1st and 3rd samples of \Cref{fig:6}), text with realistic backgrounds (as shown in the 2nd sample of \Cref{fig:4} and the 4th sample of \Cref{fig:6}), as well as natural and detailed hands, flowers, and wheel hubs (as seen in the two samples of \Cref{fig:3} and the 2nd sample of \Cref{fig:6}). In contrast, results produced by competitors exhibit excessive sharpening and diminished textural fidelity. \\
\indent \textbf{Complexity comparison.} We assess the complexity of the proposed method in comparison to SOTA diffusion-based approaches, as detailed in \Cref{tab:4}. All approaches are tested on the $\times$4 SR tasks with input images sized at 128$\times$128 pixels, and the inference time is measured on an V100 GPU. Additionally, we also report the inference time and trainable parameters of different SR methods as shown in \Cref{tab:1}. Note that the runtime is tested on the $\times$4 (64$\to$128) SR tasks with NVIDIA Telsa A100 GPU.
From \Cref{tab:1} and \Cref{tab:4}, VPD-SR significantly improves the inference speed of the teacher model by approximately tenfold, while maintaining low computational time consumption and trainable parameters compared to multi-step SD-based SR methods. It is worth emphasizing that OSEDiff employs Low-Rank Adaptation (LoRA) fine-tuning, thus requiring only a minimal number of trainable parameters.\\
\begin{table*}[!t]
    \caption{Time complexity comparison among different diffusion-based SR approaches. All approaches are tested on the $\times$4 (128$\to$512) SR tasks, and the inference time is measured on an V100 GPU.\label{tab:4}}
    \centering
    \renewcommand{\arraystretch}{1.2} 
    \renewcommand\tabcolsep{8pt}
    \begin{tabular}{c|ccccccc}
        \hline
        \multirow{2}{*}{Complexity} & \multicolumn{7}{c}{Methods} \\ 
        \cline{2-8} 
        & StableSR~\cite{20stablesr}  & DiffBIR~\cite{21diffbir} & SeeSR~\cite{22seesr}  & PASD~\cite{23pasd}  &SSP-IR~\cite{59SSPIR} &OSEDiff~\cite{16osed}  & VPD-SR \\
        \hline
        Inference Step            & 200   & 50   & 50  & 20  & 50 & 1 & 1 \\                   
        
        Inference Time (s)        & 17.76 & 11.95  & 8.40 & 13.51 &15.52 & 0.48 & 0.64\\
        Trainable Param(M)        & 153.27 & 378.94 & 751.66 & 609.52 &997.80 &8.50 & 118.59\\
        \hline
    \end{tabular}
    \vspace{-0.3cm} 
\end{table*}
\begin{table*}[!t]
    \caption{Ablation study of the proposed methods on one synthetic and two real-world datasets. * indicates that the RealSR dataset is same as that in ResShift and SinSR. The best results are highlighted in \textbf{bold}.}
    \label{tab:5}
    \centering
        \renewcommand{\arraystretch}{1.1} 
        \renewcommand\tabcolsep{8pt}
        \begin{tabular}{cccc|ccc|cc|cc}
            \hline
            \multicolumn{4}{c|}{\multirow{2}{*}{Loss Function}} & \multicolumn{7}{c}{Datasets} \\ 
            \cline{5-11} 
            & & & & \multicolumn{3}{c|}{ImageNet-Test} & \multicolumn{2}{c|}{RealSR*} 
                                                        & \multicolumn{2}{c}{RealSet65} \\ 
            \hline
            ${\cal L}_{distill}$  & ${\cal L}_{HFP}$ & ${\cal L}_{SD}$ & ${\cal L}_{adv}^{f_\theta }$  & LPIPS$\downarrow$  
            & CLIPIQA$\uparrow$  & MUSIQ$\uparrow$ & CLIPIQA$\uparrow$ 
            & MUSIQ$\uparrow$   & CLIPIQA$\uparrow$  & MUSIQ$\uparrow$         \\
            \hline
            $\checkmark$   &  &  &   & 0.2321 & 0.6125 &                                53.5767 & 0.6819 & 62.1653 & 0.6842 & 63.8796 \\
            $\checkmark$   & $\checkmark$ &  &  & 0.2278 & 0.6311 & 54.6825 &                                         0.6888 & 62.4475 & 0.6997 & 64.5043 \\
            $\checkmark$   & $\checkmark$ & $\checkmark$ & & 0.2291 & 0.6548  &                                       55.6352 & 0.7132 & 62.5075 & 0.7170 & 65.4430 \\
            $\checkmark$  & $\checkmark$ & $\checkmark$ & $\checkmark$ 
            & \textbf{0.2263} & \textbf{0.6830}  & \textbf{59.5700} & \textbf{0.7635} 
            & \textbf{63.5923} & \textbf{0.7817} & \textbf{67.9176}\\ 
            \hline
        \end{tabular}
        \vspace{-0.5cm} 
\end{table*}
\vspace{-0.5cm} 
\subsection{Ablation Study}
In the ablation study, we train VPD-SR on the training set of ImageNet, and evaluate the performance of our model on both synthetic and real-world datasets. The results of the ablation study are presented in \Cref{tab:5}, where we select no-reference metrics as the primary metrics for comparison, as these are critical for assessing image quality.\\
\textbf{Effectiveness of High-Frequency Perception Loss.} We first investigate the effectiveness and importance of high-frequency perception loss. As shown in \Cref{tab:5}, compare to the model trained solely with the vanilla distillation loss, the model trained with the vanilla distillation loss and high-frequency perception loss significantly improves the CLIPIQA and MUSIQ scores on the ImageNet-Test, RealSR, and RealSet65 datasets. This demonstrates that high-frequency information in the generated outputs plays a crucial role in strengthening semantic alignment with ground truth images, thereby validating the efficacy of our high-frequency perceptual loss.\\
\textbf{Effectiveness of Explicit Semantic-aware Supervision.} Next, we evaluate the impact of explicit semantic-aware supervision, which provides semantic supervision, on the quality of reconstructed outputs.
During training, we utilize the semantic supervision to align the GT and SR image in the image embedding space of CLIP model, to improve the semantic consistency and perception quality of generated images. The results are presented in \Cref{tab:5}. It can be observed that, in addition to the above two losses, the model also trained with semantic distillation loss, improves the CLIPIQA and MUSIQ scores across all three datasets, highlighting the role of semantic distillation in maintaining semantic consistency and perception quality of generated images. \\
\textbf{Effectiveness of Adversarial Loss.} To further enhance the authenticity of the generated results, we additionally utilize a patch-based discriminator to participate in adversarial training. 
This discriminator focuses on capturing and evaluating the authenticity of local image regions, effectively identifying and penalizing artifacts and unnatural textures in super-resolution results, guiding the generator to optimize local details that the human visual system pays more attention to.
The results in \Cref{tab:5} show that the model trained with all of the above losses achieves comprehensive improvements in CLIPIQA and MUSIQ scores across the three datasets. \\
\vspace{-0.3cm} 
\section{Conclusion and Limitation}
In this paper, we propose a visual perception diffusion distillation method that accelerates diffusion-based SR model to a single-step inference. We introduce a semantic distillation technique that utilizes explicit semantic-aware supervision extracted from the ground-truth images through the pre-trained CLIP model to enforce semantic alignment between the generated outputs and the GT images, thereby enhancing the semantic consistency and perception quality in the student model’s output. Additionally, we construct a high-frequency perception loss to preserve detailed features in SR images that contribute to the visual perception quality. Furthermore, we incorporate generative adversarial learning into the diffusion model framework to enhance the authenticity of the generated results. Extensive experiments demonstrate that our method can achieves superior performance on synthetic and real-world datasets in a single inference step.\\
\indent To be honest, although our method has a clear advantage in no-reference metrics such as CLIPIQA and MUSIQ, it shows minor limitations in full-reference fidelity metrics. Additionally, compared to multi-step SD-based SR methods, our approach requires fewer parameters, but still exceeds $100$M. We aim to design a more lightweight denoising network while addressing the trade-off between fidelity and perceptual quality in future work.

\vfill

\bibliographystyle{IEEEtran}
\bibliography{ref}

\begin{thebibliography}{10}
\providecommand{\url}[1]{#1}
\csname url@samestyle\endcsname
\providecommand{\newblock}{\relax}
\providecommand{\bibinfo}[2]{#2}
\providecommand{\BIBentrySTDinterwordspacing}{\spaceskip=0pt\relax}
\providecommand{\BIBentryALTinterwordstretchfactor}{4}
\providecommand{\BIBentryALTinterwordspacing}{\spaceskip=\fontdimen2\font plus
\BIBentryALTinterwordstretchfactor\fontdimen3\font minus \fontdimen4\font\relax}
\providecommand{\BIBforeignlanguage}[2]{{%
\expandafter\ifx\csname l@#1\endcsname\relax
\typeout{** WARNING: IEEEtran.bst: No hyphenation pattern has been}%
\typeout{** loaded for the language `#1'. Using the pattern for}%
\typeout{** the default language instead.}%
\else
\language=\csname l@#1\endcsname
\fi
#2}}
\providecommand{\BIBdecl}{\relax}
\BIBdecl

\bibitem{1survey}
Z.~Wang, J.~Chen, and S.~C. Hoi, ``Deep learning for image super-resolution: A survey,'' \emph{IEEE TPAMI}, vol.~43, no.~10, pp. 3365--3387, 2020.

\bibitem{56medical_TCSVT1}
J.~Wei, G.~Yang, W.~Wei, A.~Liu, and X.~Chen, ``Multi-contrast mri arbitrary-scale super-resolution via dynamic implicit network,'' \emph{IEEE Transactions on Circuits and Systems for Video Technology}, pp. 1--1, 2025.

\bibitem{56medical_TCSVT2}
H.~Wang, X.~Hu, X.~Zhao, and Y.~Zhang, ``Wide weighted attention multi-scale network for accurate mr image super-resolution,'' \emph{IEEE Transactions on Circuits and Systems for Video Technology}, vol.~32, no.~3, pp. 962--975, 2022.

\bibitem{57remote_TMM}
Y.~Xiao, Q.~Yuan, K.~Jiang, Y.~Chen, Q.~Zhang, and C.-W. Lin, ``Frequency-assisted mamba for remote sensing image super-resolution,'' \emph{IEEE Transactions on Multimedia}, vol.~27, pp. 1783--1796, 2025.

\bibitem{57remotesensing_TCSVT}
Y.~Xiao, Q.~Yuan, K.~Jiang, X.~Jin, J.~He, L.~Zhang, and C.-w. Lin, ``Local-global temporal difference learning for satellite video super-resolution,'' \emph{IEEE Transactions on Circuits and Systems for Video Technology}, vol.~34, no.~4, pp. 2789--2802, 2023.

\bibitem{58video_tcsvt}
D.~Li, Y.~Liu, Z.~Wang, and J.~Yang, ``Video rescaling with recurrent diffusion,'' \emph{IEEE Transactions on Circuits and Systems for Video Technology}, vol.~34, no.~10, pp. 9386--9399, 2024.

\bibitem{Lu}
S.~Ye and J.~Lu, ``Sequence unlearning for sequential recommender systems,'' in \emph{AI 2023: Advances in Artificial Intelligence}, 2024, pp. 403--415.

\bibitem{xia2022cbash}
R.~Xia, G.~Li, Z.~Huang, H.~Meng, and Y.~Pang, ``Cbash: Combined backbone and advanced selection heads with object semantic proposals for weakly supervised object detection,'' \emph{IEEE Transactions on Circuits and Systems for Video Technology}, vol.~32, no.~10, pp. 6502--6514, 2022.

\bibitem{Ye1}
S.~Ye and J.~Lu, ``Robust recommender systems with rating flip noise,'' \emph{ACM Trans. Intell. Syst. Technol.}, vol.~16, no.~1, pp. 1--19, 2024.

\bibitem{2gan}
I.~Goodfellow, J.~Pouget-Abadie, M.~Mirza, B.~Xu, D.~Warde-Farley, S.~Ozair, A.~Courville, and Y.~Bengio, ``Generative adversarial nets,'' in \emph{NeuralIPS}, 2014, pp. 2672--2680.

\bibitem{3ldl}
J.~Liang, H.~Zeng, and L.~Zhang, ``Details or artifacts: A locally discriminative learning approach to realistic image super-resolution,'' in \emph{CVPR}, 2022, pp. 5657--5666.

\bibitem{4esrgan}
X.~Wang, K.~Yu, S.~Wu, J.~Gu, Y.~Liu, C.~Dong, Y.~Qiao, and C.~Change~Loy, ``Esrgan: Enhanced super-resolution generative adversarial networks,'' in \emph{ECCVW}, 2018, pp. 63--79.

\bibitem{5realesrgan}
X.~Wang, L.~Xie, C.~Dong, and Y.~Shan, ``Real-esrgan: Training real-world blind super-resolution with pure synthetic data,'' in \emph{ICCV}, 2021, pp. 1905--1914.

\bibitem{7esrt}
Z.~Lu, J.~Li, H.~Liu, C.~Huang, L.~Zhang, and T.~Zeng, ``Transformer for single image super-resolution,'' in \emph{CVPR}, 2022, pp. 457--466.

\bibitem{8IPT}
H.~Chen, Y.~Wang, T.~Guo, C.~Xu, Y.~Deng, Z.~Liu, S.~Ma, C.~Xu, C.~Xu, and W.~Gao, ``Pre-trained image processing transformer,'' in \emph{CVPR}, 2021, pp. 12\,299--12\,310.

\bibitem{rsa}
G.~Li, J.~Shi, Y.~Zong, F.~Wang, T.~Wang, and Y.~Gong, ``Learning attention from attention: Efficient self-refinement transformer for face super-resolution.'' in \emph{IJCAI}, 2023, pp. 1035--1043.

\bibitem{freqformer}
T.~Dai, J.~Wang, H.~Guo, J.~Li, J.~Wang, and Z.~Zhu, ``Freqformer: frequency-aware transformer for lightweight image super-resolution,'' in \emph{IJCAI}, 2024, pp. 731--739.

\bibitem{10sr3}
C.~Saharia, J.~Ho, W.~Chan, T.~Salimans, D.~J. Fleet, and M.~Norouzi, ``Image super-resolution via iterative refinement,'' \emph{IEEE TPAMI}, vol.~45, no.~4, pp. 4713--4726, 2022.

\bibitem{11srdiff}
H.~Li, Y.~Yang, M.~Chang, S.~Chen, H.~Feng, Z.~Xu, Q.~Li, and Y.~Chen, ``Srdiff: Single image super-resolution with diffusion probabilistic models,'' \emph{Neurocomputing}, vol. 479, pp. 47--59, 2022.

\bibitem{12resdiff}
S.~Shang, Z.~Shan, G.~Liu, L.~Wang, X.~Wang, Z.~Zhang, and J.~Zhang, ``Resdiff: Combining cnn and diffusion model for image super-resolution,'' in \emph{AAAI}, 2024, pp. 8975--8983.

\bibitem{18come}
H.~Chung, B.~Sim, and J.~C. Ye, ``Come-closer-diffuse-faster: Accelerating conditional diffusion models for inverse problems through stochastic contraction,'' in \emph{CVPR}, 2022, pp. 12\,413--12\,422.

\bibitem{19sdedit}
C.~Meng, Y.~He, Y.~Song, J.~Song, J.~Wu, J.-Y. Zhu, and S.~Ermon, ``Sdedit: Guided image synthesis and editing with stochastic differential equations,'' in \emph{ICLR}, 2022.

\bibitem{13resshift}
Z.~Yue, J.~Wang, and C.~C. Loy, ``Resshift: Efficient diffusion model for image super-resolution by residual shifting,'' in \emph{NeurIPS}, 2023.

\bibitem{20stablesr}
J.~Wang, Z.~Yue, S.~Zhou, K.~C. Chan, and C.~C. Loy, ``Exploiting diffusion prior for real-world image super-resolution,'' \emph{IJCV}, vol. 132, no.~12, pp. 5929--5949, 2024.

\bibitem{21diffbir}
X.~Lin, J.~He, Z.~Chen, Z.~Lyu, B.~Dai, F.~Yu, Y.~Qiao, W.~Ouyang, and C.~Dong, ``Diffbir: Toward blind image restoration with generative diffusion prior,'' in \emph{ECCV}, 2024, pp. 430--448.

\bibitem{22seesr}
R.~Wu, T.~Yang, L.~Sun, Z.~Zhang, S.~Li, and L.~Zhang, ``Seesr: Towards semantics-aware real-world image super-resolution,'' in \emph{CVPR}, 2024, pp. 25\,456--25\,467.

\bibitem{23pasd}
T.~Yang, R.~Wu, P.~Ren, X.~Xie, and L.~Zhang, ``Pixel-aware stable diffusion for realistic image super-resolution and personalized stylization,'' in \emph{ECCV}, 2024, pp. 74--91.

\bibitem{26ddim}
J.~Song, C.~Meng, and S.~Ermon, ``Denoising diffusion implicit models,'' \emph{arXiv preprint arXiv:2010.02502}, 2020.

\bibitem{29dpm++}
C.~Lu, Y.~Zhou, F.~Bao, J.~Chen, C.~Li, and J.~Zhu, ``Dpm-solver++: Fast solver for guided sampling of diffusion probabilistic models,'' \emph{arXiv preprint arXiv:2211.01095}, 2022.

\bibitem{30progressive}
T.~Salimans and J.~Ho, ``Progressive distillation for fast sampling of diffusion models,'' \emph{arXiv preprint arXiv:2202.00512}, 2022.

\bibitem{32add}
A.~Sauer, D.~Lorenz, A.~Blattmann, and R.~Rombach, ``Adversarial diffusion distillation,'' in \emph{ECCV}, 2024, pp. 87--103.

\bibitem{14sinsr}
Y.~Wang, W.~Yang, X.~Chen, Y.~Wang, L.~Guo, L.-P. Chau, Z.~Liu, Y.~Qiao, A.~C. Kot, and B.~Wen, ``Sinsr: diffusion-based image super-resolution in a single step,'' in \emph{CVPR}, 2024, pp. 25\,796--25\,805.

\bibitem{15tadsr}
X.~He, H.~Tang, Z.~Tu, J.~Zhang, K.~Cheng, H.~Chen, Y.~Guo, M.~Zhu, N.~Wang, X.~Gao \emph{et~al.}, ``One step diffusion-based super-resolution with time-aware distillation,'' \emph{arXiv preprint arXiv:2408.07476}, 2024.

\bibitem{33addsr}
R.~Xie, Y.~Tai, C.~Zhao, K.~Zhang, Z.~Zhang, J.~Zhou, X.~Ye, Q.~Wang, and J.~Yang, ``Addsr: Accelerating diffusion-based blind super-resolution with adversarial diffusion distillation,'' \emph{arXiv preprint arXiv:2404.01717}, 2024.

\bibitem{16osed}
R.~Wu, L.~Sun, Z.~Ma, and L.~Zhang, ``One-step effective diffusion network for real-world image super-resolution,'' \emph{arXiv preprint arXiv:2406.08177}, 2024.

\bibitem{34s3r}
A.~Zhang, Z.~Yue, R.~Pei, W.~Ren, and X.~Cao, ``Degradation-guided one-step image super-resolution with diffusion priors,'' \emph{arXiv preprint arXiv:2409.17058}, 2024.

\bibitem{35clip}
A.~Radford, J.~W. Kim, C.~Hallacy, A.~Ramesh, G.~Goh, S.~Agarwal, G.~Sastry, A.~Askell, P.~Mishkin, J.~Clark \emph{et~al.}, ``Learning transferable visual models from natural language supervision,'' in \emph{ICML}.\hskip 1em plus 0.5em minus 0.4em\relax PMLR, 2021, pp. 8748--8763.

\bibitem{59SSPIR}
Y.~Zhang, H.~Zhang, Z.~Cheng, R.~Xie, L.~Song, and W.~Zhang, ``Ssp-ir: Semantic and structure priors for diffusion-based realistic image restoration,'' \emph{IEEE Transactions on Circuits and Systems for Video Technology}, pp. 1--1, 2025.

\bibitem{41srcnn}
C.~Dong, C.~C. Loy, K.~He, and X.~Tang, ``Learning a deep convolutional network for image super-resolution,'' in \emph{ECCV}, 2014, pp. 184--199.

\bibitem{42multi}
X.~Wu, K.~Zhang, Y.~Hu, X.~He, and X.~Gao, ``Multi-scale non-local attention network for image super-resolution,'' \emph{Signal Processing}, vol. 218, p. 109362, 2024.

\bibitem{9swinir}
J.~Liang, J.~Cao, G.~Sun, K.~Zhang, L.~Van~Gool, and R.~Timofte, ``Swinir: Image restoration using swin transformer,'' in \emph{ICCV}, 2021, pp. 1833--1844.

\bibitem{43realsrjpeg}
X.~Ji, Y.~Cao, Y.~Tai, C.~Wang, J.~Li, and F.~Huang, ``Real-world super-resolution via kernel estimation and noise injection,'' in \emph{CVPRW}, 2020, pp. 466--467.

\bibitem{44bsrgan}
K.~Zhang, J.~Liang, L.~Van~Gool, and R.~Timofte, ``Designing a practical degradation model for deep blind image super-resolution,'' in \emph{ICCV}, 2021, pp. 4791--4800.

\bibitem{45ldm}
R.~Rombach, A.~Blattmann, D.~Lorenz, P.~Esser, and B.~Ommer, ``High-resolution image synthesis with latent diffusion models,'' in \emph{CVPR}, 2022, pp. 10\,684--10\,695.

\bibitem{25iddpm}
A.~Q. Nichol and P.~Dhariwal, ``Improved denoising diffusion probabilistic models,'' in \emph{ICML}, 2021, pp. 8162--8171.

\bibitem{46ddpm}
J.~Ho, A.~Jain, and P.~Abbeel, ``Denoising diffusion probabilistic models,'' in \emph{NeurIPS}, 2020.

\bibitem{27dpmsolver}
C.~Lu, Y.~Zhou, F.~Bao, J.~Chen, C.~Li, and J.~Zhu, ``Dpm-solver: A fast ode solver for diffusion probabilistic model sampling in around 10 steps,'' in \emph{NeurIPS}, 2022.

\bibitem{47distillation}
C.~Meng, R.~Rombach, R.~Gao, D.~Kingma, S.~Ermon, J.~Ho, and T.~Salimans, ``On distillation of guided diffusion models,'' in \emph{CVPR}, 2023, pp. 14\,297--14\,306.

\bibitem{48YONOS-SR}
M.~Noroozi, I.~Hadji, B.~Martinez, A.~Bulat, and G.~Tzimiropoulos, ``You only need one step: Fast super-resolution with stable diffusion via scale distillation,'' in \emph{ECCV}, 2024, pp. 145--161.

\bibitem{31prolificdreamer}
Z.~Wang, C.~Lu, Y.~Wang, F.~Bao, C.~Li, H.~Su, and J.~Zhu, ``Prolificdreamer: High-fidelity and diverse text-to-3d generation with variational score distillation,'' in \emph{NeurIPS}, 2023.

\bibitem{49vqgan}
P.~Esser, R.~Rombach, and B.~Ommer, ``Taming transformers for high-resolution image synthesis,'' in \emph{CVPR}, 2021, pp. 12\,873--12\,883.

\bibitem{50patch}
I.~Phillip, Z.~Jun-Yan, Z.~Tinghui, A.~Alexei \emph{et~al.}, ``Image-to-image translation with conditional adversarial networks,'' in \emph{CVPR}, 2017, pp. 5967--5976.

\bibitem{54dasr}
J.~Liang, H.~Zeng, and L.~Zhang, ``Efficient and degradation-adaptive network for real-world image super-resolution,'' in \emph{ECCV}, 2022, pp. 574--591.

\bibitem{51imagenet}
J.~Deng, W.~Dong, R.~Socher, L.-J. Li, K.~Li, and L.~Fei-Fei, ``Imagenet: A large-scale hierarchical image database,'' in \emph{CVPR}.\hskip 1em plus 0.5em minus 0.4em\relax Ieee, 2009, pp. 248--255.

\bibitem{52realsr}
J.~Cai, H.~Zeng, H.~Yong, Z.~Cao, and L.~Zhang, ``Toward real-world single image super-resolution: A new benchmark and a new model,'' in \emph{ICCV}, 2019, pp. 3086--3095.

\bibitem{53drealsr}
P.~Wei, Z.~Xie, H.~Lu, Z.~Zhan, Q.~Ye, W.~Zuo, and L.~Lin, ``Component divide-and-conquer for real-world image super-resolution,'' in \emph{ECCV}, 2020, pp. 101--117.

\bibitem{40lpips}
R.~Zhang, P.~Isola, A.~A. Efros, E.~Shechtman, and O.~Wang, ``The unreasonable effectiveness of deep features as a perceptual metric,'' in \emph{CVPR}, 2018, pp. 586--595.

\bibitem{36clipiqa}
J.~Wang, K.~C. Chan, and C.~C. Loy, ``Exploring clip for assessing the look and feel of images,'' in \emph{AAAI}, 2023, pp. 2555--2563.

\bibitem{37musiq}
J.~Ke, Q.~Wang, Y.~Wang, P.~Milanfar, and F.~Yang, ``Musiq: Multi-scale image quality transformer,'' in \emph{ICCV}, 2021, pp. 5148--5157.

\bibitem{39maniqa}
S.~Yang, T.~Wu, S.~Shi, S.~Lao, Y.~Gong, M.~Cao, J.~Wang, and Y.~Yang, ``Maniqa: Multi-dimension attention network for no-reference image quality assessment,'' in \emph{CVPR}, 2022, pp. 1191--1200.

\bibitem{38niqe}
L.~Zhang, L.~Zhang, and A.~C. Bovik, ``A feature-enriched completely blind image quality evaluator,'' \emph{IEEE TIP}, vol.~24, no.~8, pp. 2579--2591, 2015.

\end{thebibliography}

\begin{IEEEbiography}[{\includegraphics[width=1in,height=1.25in,clip,keepaspectratio]{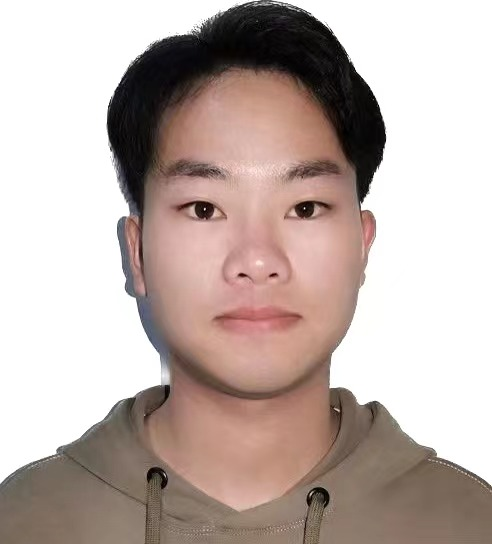}}]{Xue Wu}
 received the B.Sc. and M.Sc. degrees from Xi’an Polytechnic University, Xi’an, China, in 2021 and 2024, respectively. He is currently pursuing the Ph.D. degree in information and telecommunications engineering with Xidian University. His research interests include deep learning and image super-resolution.
\end{IEEEbiography}
\begin{IEEEbiography}[{\includegraphics[width=1in,height=1.25in,clip,keepaspectratio]{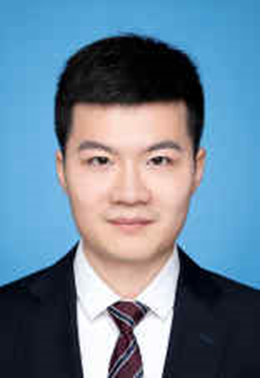}}]{Jingwei Xin}
 received the B.Sc. degree in electronic information science and technology from Shaanxi Normal University, Xi'an, China, in 2016, and the Ph.D, degree in information and telecommunications engineering from Xidian University, Xi'an, in 2021. He is currently a Lecturer with the State Key Laboratory of Integrated Services Networks, Xidian University. His research has been published in prestigious journals and academic venues, such as International Journal of Computer Vision (IJCV) and European Conference on Computer Vision (ECCV). He has broad research interests in machine learning, pattern recognition, and computer vision.
\end{IEEEbiography}
\begin{IEEEbiography}[{\includegraphics[width=1in,height=1.25in,clip,keepaspectratio]{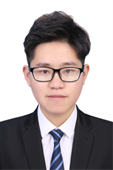}}]{Zhijun Tu}
 received both his B.Sc. and M.Sc. degrees in 2019 and 2022, respectively, from Xi’an Jiaotong University, Xi’an, China. Since 2022, he has been a Researcher at Huawei Noah’s Ark Lab, Shanghai, China. His research has been published in top-tier artificial intelligence (AI) conferences, such as the European Conference on Computer Vision (ECCV), IEEE/CVF Conference on Computer Vision and Pattern Recognition (CVPR), and Conference on Neural Information Processing Systems (NeurIPS) with a focus on lightweight and efficient large language models (LLMs) and AIGC (AI-generated content) systems.
\end{IEEEbiography}
\begin{IEEEbiography}[{\includegraphics[width=1in,height=1.25in,clip,keepaspectratio]{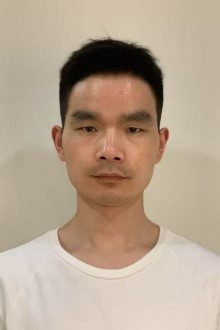}}]{Jie Hu}
 received the B.Sc. degree from AnHui University, HeFei, China, in 2013, and the M.Sc. degrees from University of Science and Technology of China, HeFei, in 2016, respectively. He is currently a Researcher with HUAWEI Company. His research has been published in prestigious journals and academic venues, such as IEEE TRANSACTIONS ON IMAGE PROCESSING (TIP) and European Conference on Computer Vision (ECCV). His research interests include low-level vision and AIGC.
\end{IEEEbiography}
\begin{IEEEbiography}[{\includegraphics[width=1in,height=1.25in,clip,keepaspectratio]{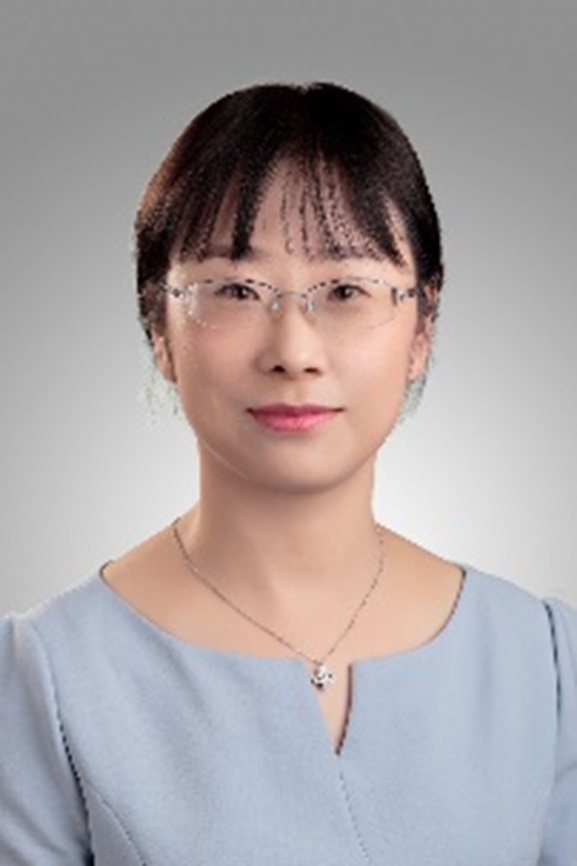}}]{Jie Li}
 received the B.Sc. degree in electronic engineering, the M.Sc. degree in signal and information processing, and the Ph.D. degree in circuit and systems from Xidian University, Xi’an, China, in 1995, 1998, and 2004, respectively. She is currently a Professor with the School of Electronic Engineering, Xidian University. Her research interests include image processing and machine learning. In these areas, she has published around 50 technical articles in refereed journals and proceedings, including IEEE TRANSACTIONS ON NEURAL NETWORKS AND LEARNING SYSTEMS, IEEE TRANSACTIONS ON IMAGE PROCESSING, IEEE TRANSACTIONS ON CIRCUITS AND SYSTEMS FOR VIDEO TECHNOLOGY, and Information Sciences.
\end{IEEEbiography}
\begin{IEEEbiography}[{\includegraphics[width=1in,height=1.25in,clip,keepaspectratio]{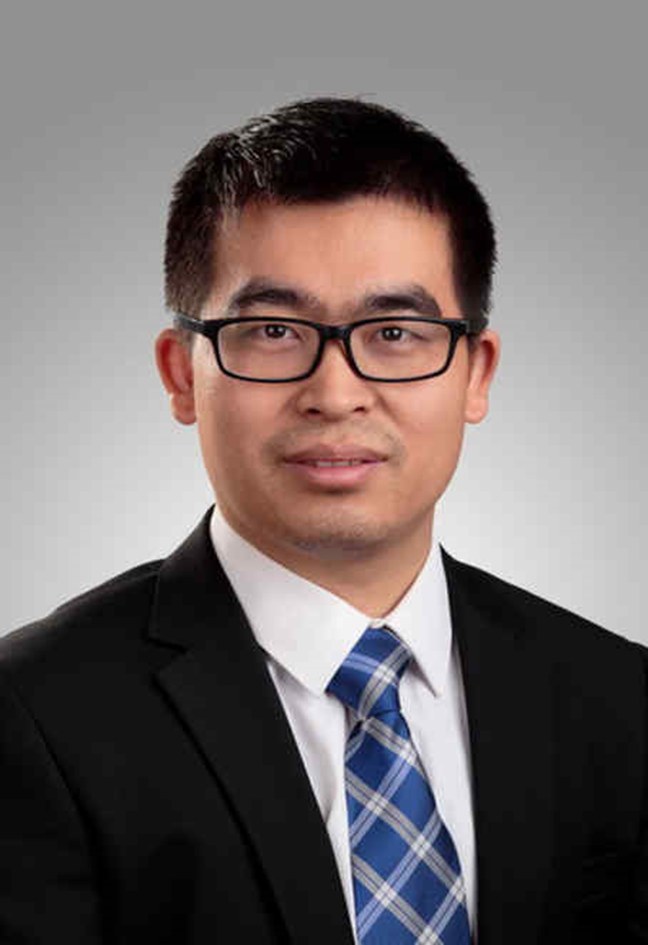}}]{Nannan Wang}
 (Senior Member, IEEE) received the B.Sc. degree in information and computation science from the Xi’an University of Posts and Telecommunications in 2009 and the Ph.D. degree in information and telecommunications engineering from Xidian University in 2015. He is currently a Professor with the State Key Laboratory of Integrated Services Networks, Xidian University. He has published over 150 articles in refereed journals and proceedings, including IEEE TRANSACTIONS ON PATTERN ANALYSIS AND MACHINE INTELLIGENCE, International Journal of Computer Vision, CVPR, and ICCV. His current research interests include computer vision and machine learning.
\end{IEEEbiography}
\begin{IEEEbiography}[{\includegraphics[width=1in,height=1.25in,clip,keepaspectratio]{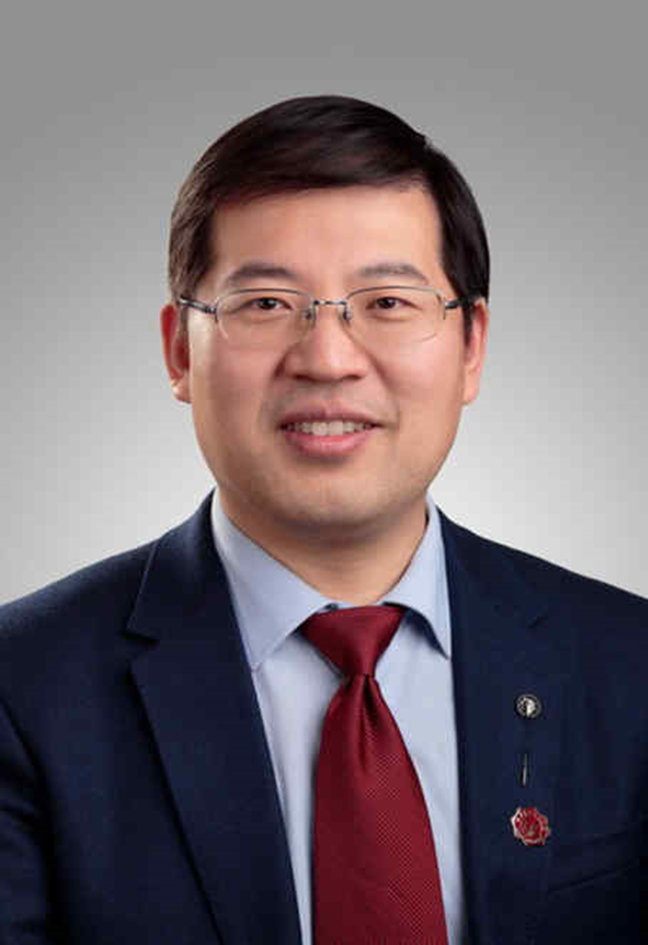}}]{Xinbo Gao}
 (Fellow, IEEE) received the B.Eng., M.Sc. and Ph.D. degrees in electronic engineering, signal and information processing from Xidian University, Xi’an, China, in 1994, 1997, and 1999, respectively. From 1997 to 1998, he was a research fellow at the Department of Computer Science, Shizuoka University, Shizuoka, Japan. From 2000 to 2001, he was a post-doctoral research fellow at the Department of Information Engineering, the Chinese University of Hong Kong, Hong Kong. Since 1999, he has been at the School of Electronic Engineering, Xidian University and now he is a Professor of Pattern Recognition and Intelligent System of Xidian University. Since 2020, he has been also a Professor of Computer Science and Technology of Chongqing University of Posts and Telecommunications. His current research interests include computer vision, machine learning and pattern recognition. He has published seven books and around 300 technical articles in refereed journals and proceedings. Prof. Gao is on the Editorial Boards of several journals, including Signal Processing (Elsevier) and Neurocomputing (Elsevier). He served as the General Chair/Co-Chair, Program Committee Chair/Co-Chair, or PC Member for around 30 major international conferences. He is Fellows of IEEE, IET, AAIA, CIE, CCF, and CAAI.
\end{IEEEbiography}


\end{document}